\documentclass[journal]{vgtc}                % final (journal style)
\ifpdf%                                % if we use pdflatex
  \pdfoutput=1\relax                   % create PDFs from pdfLaTeX
  \usepackage{graphicx}                % allow us to embed graphics files
  \DeclareGraphicsExtensions{.pdf,.png,.jpg,.jpeg} % for pdflatex we expect .pdf, .png, or .jpg files
\else%                                 % else we use pure latex
  \usepackage{graphicx}                % allow us to embed graphics files
  \DeclareGraphicsExtensions{.eps}     % for pure latex we expect eps files
\fi%

\graphicspath{{figures/}{pictures/}{images/}{./}} % where to search for the images

\usepackage{microtype}                 % use micro-typography (slightly more compact, better to read)
\PassOptionsToPackage{warn}{textcomp}  % to address font issues with \textrightarrow
\usepackage{textcomp}                  % use better special symbols
\usepackage{mathptmx}                  % use matching math font
\usepackage{times}                     % we use Times as the main font
\usepackage{cite}                      % needed to automatically sort the references
\usepackage{tabu}                      % only used for the table example
\usepackage{booktabs}                  % only used for the table example
\usepackage[utf8]{inputenc} % allow utf-8 input
\usepackage[T1]{fontenc}    % use 8-bit T1 fonts
\usepackage{hyperref}       % hyperlinks
\usepackage{url}            % simple URL typesetting
\usepackage{amsfonts}       % blackboard math symbols
\usepackage{nicefrac}       % compact symbols for 1/2, etc.
\usepackage{microtype}      % microtypography
\usepackage{xcolor}         % colors
\usepackage{amsmath}
\usepackage{amssymb}
\usepackage{array}
\usepackage{arydshln}
\usepackage{multirow}
\usepackage{txfonts}
\usepackage{subfigure}
\usepackage{float}
\usepackage{wrapfig}
\usepackage{comment}
\usepackage{caption}

\definecolor{c1}{HTML}{d73027}
\definecolor{c2}{HTML}{fc8d59}
\definecolor{c3}{HTML}{998ec3}
\definecolor{c4}{HTML}{c51b7d}
\definecolor{c5}{HTML}{1a9850}
\definecolor{c6}{HTML}{01665e}
\definecolor{c7}{HTML}{4575b4}
\usepackage{tikz,pgfplots}
\usepackage{filecontents}

\ieeedoi{10.1109/TVCG.2023.3247088}

\onlineid{0}

\vgtccategory{Research}
\vgtcpapertype{please specify}

\title{LiDAR-aid Inertial Poser: Large-scale Human Motion Capture by Sparse Inertial and LiDAR Sensors}

\author{Yiming Ren$^\dagger$, Chengfeng Zhao$^\dagger$, Yannan He, Peishan Cong, Han Liang, Jingyi Yu, Lan Xu$^\star$, and Yuexin Ma$^\star$}
\authorfooter{
\item
$\dagger$: Equal contribution.
\item
$\star$: Corresponding author.
\item
 Yiming Ren, Chengfeng Zhao, Yannan He, Peishan Cong, Han Liang are with ShanghaiTech University. E-mail: \{ renym2022 | zhaochf2022 | heyn | congpsh | lianghan | yujingyi | xulan1 | mayuexin\}@shanghaitech.edu.cn.
 \item
 Jingyi Yu, Lan Xu, and Yuexin Ma are with ShanghaiTech University and Shanghai Engineering Research Center of Intelligent Vision and Imaging. E-mail: \{yujingyi | xulan1 | mayuexin\}@shanghaitech.edu.cn.
}

\shortauthortitle{Yiming Ren, Chengfeng Zhao \MakeLowercase{\textit{et al.}}: LiDAR-aid Inertial Poser: Large-scale Human Motion Capture by Sparse Inertial and LiDAR Sensors}
\abstract{
We propose a multi-sensor fusion method for capturing challenging 3D human motions with accurate consecutive local poses and global trajectories in large-scale scenarios, only using single LiDAR and 4 IMUs, which are set up conveniently and worn lightly. 
Specifically, to fully utilize the global geometry information captured by LiDAR and local dynamic motions captured by IMUs, we design a two-stage pose estimator in a coarse-to-fine manner, where point clouds provide the coarse body shape and IMU measurements optimize the local actions. 
Furthermore, considering the translation deviation caused by the view-dependent partial point cloud, we propose a pose-guided translation corrector.
It predicts the offset between captured points and the real root locations, which makes the consecutive movements and trajectories more precise and natural. Moreover, we collect a LiDAR-IMU multi-modal mocap dataset, LIPD, with diverse human actions in long-range scenarios.
Extensive quantitative and qualitative experiments on LIPD and other open datasets all demonstrate the capability of our approach for compelling motion capture in large-scale scenarios, which outperforms other methods by an obvious margin. We will release our code and captured dataset to stimulate future research.
}
\keywords{Human motion capture, shape modeling, virtual reality, sensor fusion}

\CCScatlist{ % not used in journal version
 \CCScat{K.6.1}{Management of Computing and Information Systems}%
{Project and People Management}{Life Cycle};
 \CCScat{K.7.m}{The Computing Profession}{Miscellaneous}{Ethics}
}

\teaser{
  \centering
  \includegraphics[width=\linewidth]{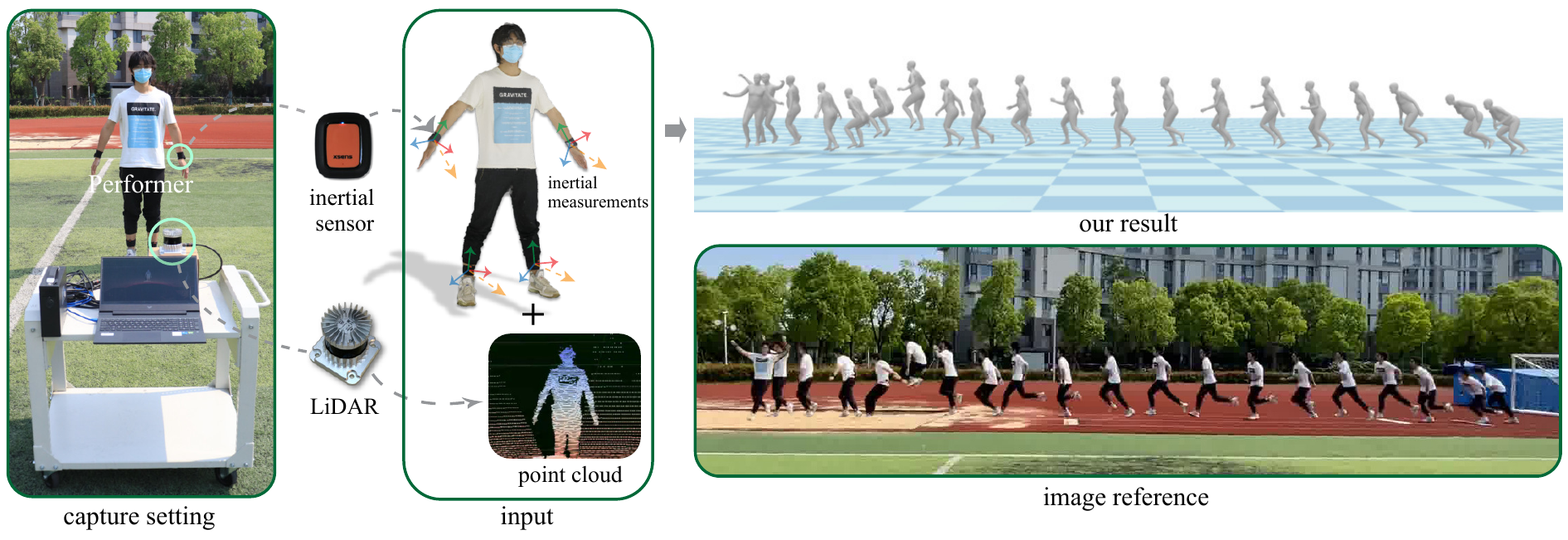}
  \caption{We propose a multi-modal motion capture approach, LIP, that estimates challenging human motions with accurate local pose and global translation in large-scale scenarios using single LiDAR and four inertial measurement units.}
  \label{fig:teaser}
}

\vgtcinsertpkg

\begin{document}


\firstsection{Introduction}

\maketitle

%% Intro
\label{sec:introduction}
With the rise of VR/AR over the last decades, human motion capture (mocap) evolves as a cutting-edge technique for humans to interact and communicate with each other in virtual worlds using our body language. Despite the huge progress of data-driven mocap solutions, the accurate and convenient capture of human motions in large-scale scenarios remains challenging. It is critical for the reconstruction, simulation, and generation of sporting mega-events, stage performances, interactions of crowds, etc., and has recently received substantive attention.

% challenge
%\textbf{The Challenges of Current Approaches:}
% vision-based method
By far, optical-based solutions take the majority of human mocap. The high-end marker-based solutions~\cite{VICON,Vlasic2007,optitrack} require outside-in multi-camera setup or dense optical markers, and thus confine the performers to a constrained captured area, making large-scale capture impractical. Recently, learning-based monocular methods~\cite{HMR18,Kanazawa_2019CVPR,VIBE_CVPR2020,zanfir2021neural,DeepCap_CVPR2020,challencap,PHMR_ICCV2021,HUMOR_ICCV2021,PARE_ICCV2021} enable robust motion capture under light-weight setting. Although alleviating expensive facilities and fixed captured region, they remain vulnerable to occlusions, lack of textures and severe changes of environment lighting, etc,. Moreover, their inherent lack of depth measurement makes them unstable to accurately track global trajectories of humans, especially under large-scale settings. 

% Inertial method
In contrast, motion capture using inertia information recorded by Inertial Measurement Units (IMUs) is occlusion-unaware and environment-independent. The high-end solutions~\cite{XSENS,noitom} require a large amount of body-worn IMUs (from 8 to 17), making them unsuitable to capture human motions with everyday apparel. Recent data-driven advances~\cite{huang2018DIP,yi2021transpose,PIPCVPR2022} enable real-time motion capture with sparse IMUs. They can obtain global location by accumulating the IMU observation and even partially alleviate the drifting with foot-ground contact or physical constraints. Yet, such purely inertial methods still inherently suffer from temporally cumulative error of global localization, especially for capturing large-scale motions with a long duration. 
% LiDAR-based method
Only recently, LiDAR-based mocap is gaining increasing attention due to the tremendous progress of LiDAR in large-scale perception~\cite{zhu2021cylindrical,zhu2020ssn,cong2022stcrowd,dai2022hsc4d}. Notably, LiDARCap~\cite{li2022lidarcap} leverages a graph-based convolutional network to predict daily human poses from the point clouds captured by a LiDAR sensor within range of 30 meters. However, it's fragile to capture challenging human motions or multi-person scenes with strong self-occlusions and external occlusions due to the view-dependent and sparsity-varying properties of LiDAR point clouds.

%\begin{figure}[t]
%\centering
%\includegraphics[width=\linewidth]{teaser.pdf}
%	\caption{We propose a multi-modal motion capture approach, LIP, that estimates challenging human motions with accurate local pose and global translation in large-scale scenarios using single LiDAR and four inertial measurement units.}
%	\label{fig:teaser}
%\vspace{-3ex}
%\end{figure}

% solution
To tackle the above challenges, we propose a novel approach called LiDAR-aid Inertial Poser (LIP), to capture challenging human motions in large-scale scenarios accurately. In stark contrast with previous mocap system, our LIP adopts a novel and light-weight hardware setup using only single LiDAR and 4 IMUs. These two kinds of sensors are complementary to each other, providing both global geometry and local dynamics information of the captured performer. Meanwhile, they are naturally privacy-preserving and insensitive to the lighting, which is appropriate to be generalized to novel scenes. As shown in \autoref{fig:teaser}, our system faithfully reconstructs both local skeletal poses and global trajectories of the performer under a novel sensor-fusion setting.

Generating robust mocap results using such novel multi-modal inputs is non-trivial. First, point cloud only capture the global visual information, while IMU measurements encode the local actions physically. Furthermore, the partially captured point clouds of human body will significantly change under various poses and views of the performer, which negatively affects the accurate localization and naturally consecutive motion capture.
% stage1: global temporal pose-prior distillation
To this end, in LIP, we adopt two consistent technical modules to accurately estimate human poses and translations, respectively. For the former one, we introduce a \textbf{Multi-modal Pose Estimation} module, which is two-stage, conducting sensor fusion and local body motion estimation in a coarse-to-fine manner. Specifically, we first extract pose and motion patterns from point cloud by regressing 24 joints of the parametric human model SMPL~\cite{SMPL2015} and the 6D rotation of the root joint through a convolutional temporal encoder (PointNet-GRU) in Stage1.
% stage2: hierarchical pose estimation
Next, a hierarchical pose estimation process with fused IMU measurements is appended to refine the coarse skeleton by a joint-map estimator and solve the inverse kinematics(IK) problem for the joint rotations by a body-pose estimator in Stage2. 
%We organize this module as 2 estimators: 1) \textit{Joint-map estimator} refines the positions of all 24 joints from the pose-prior and IMU signals; 2) \textit{Body-pose estimator} solves the inverse kinematics(IK) problem for the joint rotations from joint positions.
%
% stage3: pose-guided translation correction
To precisely capture the 3D global motion trajectories, we introduce the second module called \textbf{Pose-guided Translation Correction}, so as to learn the deviation between the partially captured point could and the real location of the root joint. Specifically, considering the fact that diverse poses can affect the deviation, we use another isomorphic PointNet-GRU structure to model the shape and pose characteristics from point cloud, cooperated with the estimated joint rotations and refined skeleton. We further utilize the fused pose feature to eliminate the inherent translation deviation and random noise from the captured point cloud. 

In particular, to facilitate the research of multi-modal human mocap tasks, we collect a huge LiDAR-IMU hybrid mocap dataset, LIPD, consisting of rich and challenging single-person and multi-person actions in long-range scenes. It is the first dataset to provide multi-modal LiDAR point cloud and IMU measurements and ground-truth SMPL pose parameters for the mocap task. 
%performance
We conduct extensive experiments on LIPD and a variety of other real and synthetic datasets~\cite{huang2018DIP,li2021ai,li2022lidarcap,AMASS_ICCV2019} to demonstrate the capability of our method LIP for capturing challenging human motions in various large-scale scenes. 
To summarize, our main contributions include:

\begin{itemize}
\item We propose the first LiDAR-IMU hybrid approach for 3D human motion capture in large-scale scenarios and achieves state-of-the-art performance.
\item We propose an effective two-stage coarse-to-fine fusion method to fully utilize the complementary features of multi-modal input for accurate pose estimation.
\item We propose an approach to eliminate the translation deviation in a pose-guided manner, achieving accurate global trajectories and natural consecutive actions.
\item We provide a huge LiDAR-IMU-based multi-modal mocap dataset with diverse human actions in various large-scale scenarios, which can benefit both single modal and multi-modal mocap research works.

\end{itemize}
\section{Related Work}

\label{sec:related work}

\textbf{Optical Motion Capture.}
% marker-based
Marker-based motion capture studios~\cite{VICON,Vlasic2007,optitrack} enable capturing high-quality professional motions, which have achieved success and are widely used in the industry. However, these systems are costly and cumbersome, and performers usually need to wear the marker suits, which means unavailable for daily usage. 
% markerless
To overcome these problems, the exploration of markerless mocap~\cite{BreglM1998,de2008performance,TheobASST2010,FlyCap,HolteTTM2012,SigalBB2010,SigalIHB2012,StollHGST2011,JooLTGNMKNS2015,UnstructureLan,yuan2020residual,luo2021dynamics} has made great progress. Previous multi-view algorithms~\cite{AminARS2009,BurenSC2013,ElhayAJTPABST2015,RhodiRRST2015,Robertini:2016,Pavlakos17,Simon17} demonstrate robust motion capture even in the wild. Although the cost and intrusiveness is drastically reduced, synchronizing and calibrating multi-camera systems is still tedious.
Thus various monocular mocap approaches have been proposed, which estimate 3D human pose and shape by optimizing~\cite{TAM_3DV2017,Lassner17,keepitSMPL,Kolotouros_2019_CVPR} or directly regressing~\cite{HMR18,Kanazawa_2019CVPR,VIBE_CVPR2020,zanfir2021neural}. To overcome various flaws of monocular setup, template-based approaches~\cite{MonoPerfCap,LiveCap2019tog,EventCap_CVPR2020,DeepCap_CVPR2020,challencap}, probabilistic approaches~\cite{PHMR_ICCV2021,HUMOR_ICCV2021} and semantic-modeling approaches~\cite{PARE_ICCV2021} are proposed. However, the inherent depth ambiguity is still unsolved.
Some approaches~\cite{Shotton:2011,Baak:2011,Wei:2012,DoubleFusion,guoTwinFusion} using the commodity depth cameras enable alleviating this, but these active IR-based cameras are unsuitable for outdoor capture and the capture volume is limited.
Recently, Li et al.~\cite{li2022lidarcap} employs a consumer-level LiDAR which provides large-scale depth information, to enable large-scale 3D human motion capture. However, LiDAR point cloud is view-dependent and sparsity-varying and the pure LiDAR-based method suffers from severe occlusions, which hinders the generalization to more challenging human actions and multi-person scenarios where occlusions are inevitable.

\textbf{Inertial Motion Capture.}
In contrast to optical approaches based on cameras, purely inertial approaches using IMUs are free from occlusion and restricted recording environment and volume. However, commercial purely inertial solutions such as Xsens MVN~\cite{XSENS} rely on large amounts of IMUs. Performers are usually required to wear tight suits with densely bounded IMUs, which is intrusive, tedious and inconvenient, and prompt the community forward to the sparse setup. 
A pioneering work, SIP~\cite{von2017SIP}, presents an optimization-based offline method using only 6 IMUs and achieves promising results. 
Inspired by it, recent data-driven approaches~\cite{huang2018DIP,yi2021transpose,PIPCVPR2022} with the same setup achieve great improvements in accuracy and efficiency, which enable real-time pose and translation estimation. 
Nevertheless, these purely inertial approaches still suffer from substantial drifts while performing high-speed or large-scale capture.

\textbf{Hybrid Motion Capture.}
As optical and inertial mocap solutions suffer from occlusions and drifts, respectively. Approaches that fuse these two types of sensors to benefit from complementarity, so as to achieve more robust mocap, have attracted much attention. Preceding methods propose to combine IMUs with RGB cameras, which can be achieved by either optimization~\cite{pons2010multisensor,kaichi2020resolving,malleson2017real,malleson2019real,malleson2020real} or regression~\cite{gilbert2019fusing,trumble2017total,von2016human,zhang2020fusing}. Recently, Liang et al.~\cite{liang2022hybridcap} presents a learning-and-optimization method fusing a single camera with only 4 IMUs, which demonstrates robust challenging motion capture. Besides, some works fuse IMUs with depth cameras~\cite{helten2013real,Zheng2018HybridFusion} or optical markers~\cite{andrews2016real}, which achieve promising results. To improve interactive and immersive experiences in AR/VR applications, motion tracking methods using head mounted devices(HMD) mixed with other sparse sensors bring great advances~\cite{jiang2022avatarposer,winkler2022questsim}. Nevertheless, these methods still suffer from limited capture volume and are sensitive to light, which limits the usage for large-scale motion capture. Furthermore, following traditional optimization schemes, several methods~\cite{patil2020fusion,ziegler2011accurate} combine dense IMUs with multiple LiDARs, which is inconvenient to set up and too heavy for performers.
In this work, we propose a novel data-driven hierarchical mocap approach by fusing only single LiDAR and 4 IMUs, which is lightweight and alleviates both occlusion and drift problems and achieves accurate 3D challenging motion capture in large-scale scenes.

\textbf{Motion Capture Dataset.}
Current booming data-driven mocap methods rely on huge labeled datasets. There are already many open datasets for facilitating related research. Human3.6M~\cite{ionescu2013human3} is a popular and wildly used motion capture dataset, which records 3.6 million frames data by 10 high speed motion cameras, it covers 17 daily motions performed by 11 actors, and a similar dataset HumanEva~\cite{SigalBB2010} provides 6 common actions. Both of them are marker-based datasets and rely on surrounding multi-view cameras to gather 3D human poses, thus the activity space, clothing, actions and lighting conditions are limited. AMASS~\cite{AMASS_ICCV2019} integrates them into a single meta dataset with a common framework and parameterization for the training of data-driven networks. MPI-INF-3DHP~\cite{mehta2017monocular} captures ground-truth from commercial marker-less motion capture in a multi-camera green screen studio and AIST++~\cite{li2021ai} captures ground-truth by 3D reconstruction method from multi-view videos. Such marker-less datasets cover more complex poses than marker-based ones. TotalCapture~\cite{trumble2017total} is the first dataset to utilize IMU to capture the pose, which consists of videos and corresponding 13 IMU sensor measurements. All above datasets are limited to indoor scenes. As the demand of mocap in outdoor large-scale scenes grows, some outdoor datasets emerged. PedX~\cite{kim2019pedx} is a large-scale outdoor 3D datasets captured by stereo cameras and LiDAR, but the full 3D labels are generated by 2D annotations. 3DPW~\cite{von2018recovering} gathers motion ground-truth by hand-held smartphone camera and IMU sensors, and DIP-IMU~\cite{huang2018DIP} uses 17 IMU sensors to generate ground-truth by SIP. Nevertheless, both camera and IMUs can not provide accurate depth information, making these methods difficult to reconstruct the human motions in the whole scene. LiDARHuman26M~\cite{li2022lidarcap} proposes the first large-scale benchmark dataset featuring LiDAR point cloud and IMU-captured human motion ground truth. It utilizes LiDAR point cloud as pure input to acquire the global localization information and eliminate the local pose ambiguity. However, the dataset does not provide raw IMU measurements and only consists of 20 simple daily single-person motions with few self-occlusion cases. We propose a LiDAR and IMU multi-sensor-based mocap dataset with diverse challenge human actions in various large-scale scenarios, which is applicable for both general and professional scenarios.

\section{Method}
\label{sec:method}
% math definitions
\newcommand{\sensor}[1]{s_{#1}}
\newcommand{\bone}[1]{b_{#1}}
\newcommand{\rot}[2]{\mathbf{R}_{#1}^{#2}}
\newcommand{\rotsixD}[2]{\mathbf{\Theta}_{#1}^{#2}}
\newcommand{\acc}[2]{\mathbf{a}_{#1}^{#2}}
\newcommand{\inertialframe}{\mathcal{F}_I}
\newcommand{\lidarframe}{\mathcal{F}_L}
\newcommand{\rootframe}{\mathcal{F}_M}
\newcommand{\localframe}[1]{\mathcal{F}_{#1}}

\label{sec:overview}
\begin{figure*}[ht!]
	\centering
	\includegraphics[width=\linewidth]{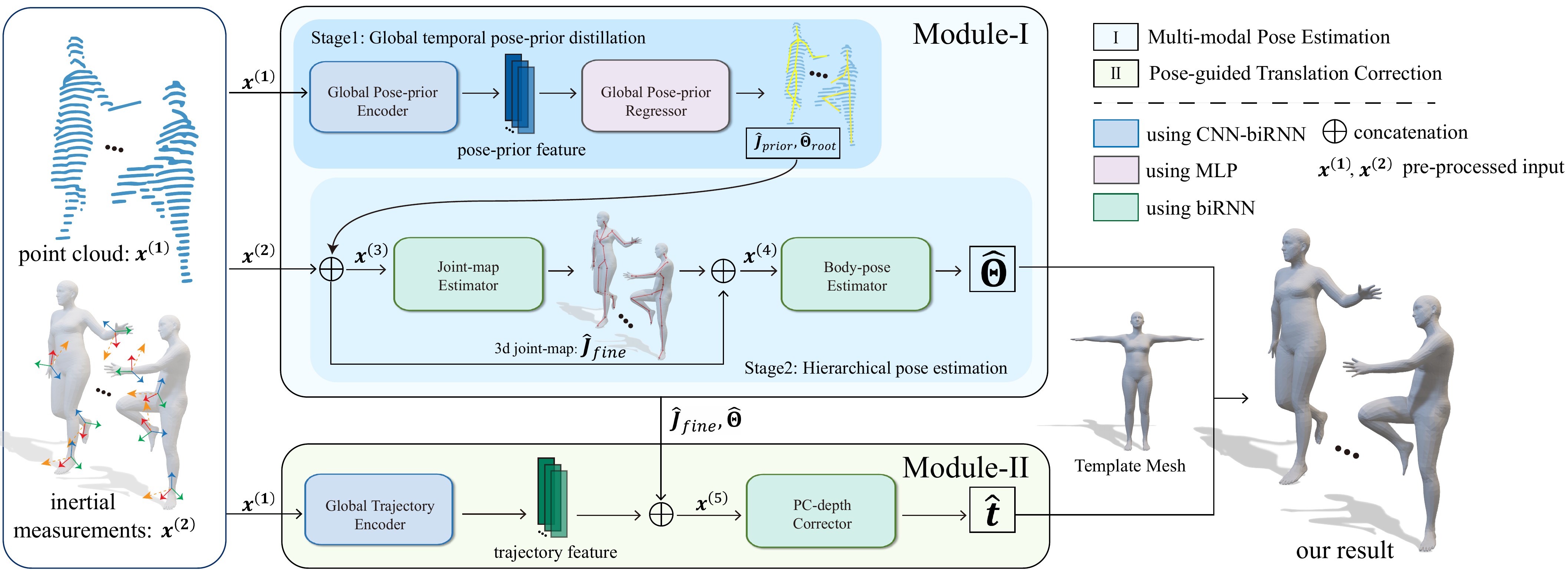}
	\caption{Overview of our pipeline. It consists of two cooperative modules: \textit{Multi-modal Pose Estimation}(Module-I) and \textit{Pose-guided Translation Correction}(Module-II) to estimate skeletal pose and global translation from sparse IMU and LiDAR inputs.}
	\label{fig:pipeline}
	\vspace{-2ex}
\end{figure*}
Our goal is to capture challenging 3D human motions in large-scale scenarios with consistently accurate local pose and global trajectory estimation using single LiDAR and 4 IMUs. \autoref{fig:pipeline} illustrates an overview of the entire pipeline, which consists of two cooperative modules to infer pose and translation, respectively. For pose inference module, we propose a two-stage network working in a coarse-to-fine manner to distill rough human body skeleton and global orientation from raw point cloud first, and then regress the human body pose from joint positions refined by IMU measurements (\autoref{sec:m1}). For translation inference module, we design a pose-guided approach to learn the latent domain gap between LiDAR measurements and real global movements, through which the inherent translation deviation and random noise from point cloud can be eliminated (\autoref{sec:m2}).

\subsection{Preliminaries}
In this section, we give detailed explanations for the design of our network. Before that, we clarify our system input pre-processing and the most frequently used math symbols in the following.

\textbf{System Input Pre-processing:} Since raw point cloud sequence with variable $N_p(t)$ points at different time frame $t$ is temporally inconsistent, we normalize the point cloud in every single frame $\widetilde{\mathbf{x}}^{(1)}(t)\in \mathbb{R}^{N_p(t)\times 3}$ to $\mathbf{x}^{(1)}(t)\in \mathbb{R}^{N_{fps}\times 3}$ by subtracting its arithmetic mean and resampling to fixed $N_{fps}$ points using farthest point sampling algorithm (FPS). In our implementation, we set $N_{fps} = 256$. For IMU measurements, we transform sensors' inertia in inertial coordinate system to bones' inertia in LiDAR coordinate system, and formulate single-frame inertial input as $\mathbf{x}^{(2)}(t) = [\rot{lw}{},\rot{rw}{},\rot{la}{},\rot{ra}{},\acc{lw}{},\acc{rw}{}\acc{la}{},\acc{ra}{}]\in \mathbb{R}^{48}$, where $\rot{}{}$ denotes the rotation in flattened rotation matrix form while $\acc{}{}$ indicates the free acceleration, and $lw, rw, la, ra$ mean left wrist, right wrist, left ankle, and right ankle, respectively.

\textbf{Definition of Motion Representations:} We define $N_j,\ N_s$ as the amount of body joints and IMU sensors; $\hat{\mathbf{J}}(t),\ \mathbf{J}^{GT}(t),\ \widetilde{\mathbf{J}}^{GT}(t)\in \mathbb{R}^{3N_j}$ as predicted root-relative joint positions, ground-truth root-relative joint positions, and ground-truth absolute joint positions at time $t$; $\hat{\mathbf{\Theta}}(t),\ \mathbf{\Theta}^{GT}(t)\in \mathbb{R}^{6N_j}$ as predicted joint rotations and ground-truth joint rotations in 6D rotation representation~\cite{zhou2019continuity} at time frame $t$. Note that all the formulations of loss functions defined below omit a common factor $\frac{1}{T}$ where $T$ is the total time length of training motion sequences.

%\begin{figure}[h]
%    \centering
%    \includegraphics[width=\linewidth]{detailed}
%    \caption{Detailed architecture of \textit{Encoder} Block, consisting of PointNet and bidirectional GRU.}
%    \label{fig:detailed_architecture}
%\end{figure}

\label{sec:dataset}
\subsection{Multi-modal Pose Estimation}
\label{sec:m1}
% instruction and modification %
Purely inertial or LiDAR-based methods more or less suffer from insufficient observations. First, without global visual cues, the reconstruction from local inertia to the 3D joint position is ambiguous and the global location is also inaccurate. Secondly, the view-dependent LiDAR point cloud lacks the representation of unseen body parts, causing difficulties in motion prediction when severe occlusions occur. Therefore, we propose to estimate body pose with multi-modal input, which includes both global geometry information from the LiDAR point cloud and local dynamic motion information from inertia measurements. However, since point cloud is in spatial form while IMU measurement is a physical quantity, the huge domain gap makes it irrational to directly concatenate $\mathbf{x}^{(1)}(t)$ and $\mathbf{x}^{(2)}(t)$ as network input directly. Instead, we formulate this module as a two-stage network working in a coarse-to-fine manner so that the point cloud can combine with motion inertia efficiently. \textit{Global Temporal Pose-prior Distillation} is designed to extract hidden geometric feature and motion pattern from the normalized point cloud and express it explicitly. After that, \textit{Hierarchical Pose Estimation} fuses IMU measurements in and regresses body joint rotations from refined 3D body joint positions. 

\textbf{Global Temporal Pose-prior Distillation:} 
Considering that the raw point cloud can directly represent the coarse human shape and pose information and sparse IMUs can not directly estimate the root joint orientation, we distill the human-skeleton-joint positions and the root orientation from the point cloud in the first stage and then use the four IMU sensors in the arms and legs to refine the result and regress the shape parameters. 
Specifically, We propose a global temporal pose-prior distillation to regress the 24 SMPL joint positions and the 6D root orientation, which is composed of a global feature extractor PointNet~\cite{qi2017pointnet} and a temporal encoder two-way GRU(bi-GRU).
PointNet is used as encoder to extract human skeleton geometric information from the raw point cloud as a feature vector $v{(t)} \in \mathbb{R}^k$ from each frame $F{(t)}$, where k = 1024.
%The encoder is in charge of summarizing the human skeleton geometric information in the raw point cloud as a feature vector $v{(t)} \in \mathbb{R}^k$ from each frame $F{(t)}$ where k = 1024, an classical point cloud extractor such as PointNet can be used to extract information efficiently and globally in the given sparse point cloud. 
We feed the frame-wise features $v{(t)}$ into the two-way GRU(bi-GRU) to generate the hidden variables $h{(t)}$ to extract temporal information. Then, we use the MLP decoder to predict the joint positions $\hat{\mathbf{J}}_{prior}{(t)}$ and the root orientation  $\hat{\mathbf{\Theta}}_{root}{(t)}$, which forms part of the second stage of input. We extract 24 SMPL body joints $\mathbf{J}^{GT}{(t)}$ and select 6D root rotation from SMPL pose parameters $\mathbf{\Theta}_{root}^{GT}{(t)}$ as the ground truth. The losses of the above two supervision information can be formulated as
\begin{equation}
    \begin{aligned}
    \mathcal{L}_{joint-prior}= \sum_t \parallel \hat{\mathbf{J}}_{prior}(t)-\mathbf{J}^{GT}(t) \parallel_2^2,
    \end{aligned}
\label{equ:loss_Joint}
\end{equation}

\begin{equation}
    \begin{aligned}
    \mathcal{L}_{ori-prior}= \sum_t \parallel\hat{\mathbf{\Theta}}_{root}(t)-\mathbf{\Theta}_{root}^{GT}(t)\parallel_2^2,
    \end{aligned}
\label{equ:globalR_loss}
\end{equation}

\begin{equation}
    \begin{aligned}
    \mathcal{L}_{prior} = \lambda_1\mathcal{L}_{joint-prior} + \lambda_2\mathcal{L}_{ori-prior},
    \end{aligned}
\end{equation}
where $\lambda_1$ and $\lambda_2$ are hyper-parameters.

\textbf{Hierarchical Pose Estimation:} 
% the pose-prior from point cloud is coarse
Although the explicit pose-prior distilled from the point cloud is available for a neural Inverse Kinematics (IK) solver, it is too coarse to give accurate 3D joint positions and robust inference for challenging motions due to the lack of locally precise motion observations. %
% our design
Therefore, we hierarchically organize this stage as two estimators, which refine the root-relative 3D joint positions with the aid of IMU measurements and then regress the joint rotations. These two estimators share the same architecture, composed of three GRUs connected sequentially with a skip connection, which can make use of temporal motion information and maintain a healthy back propagation for our deep network during training. The former one, \textit{Joint-map Estimator}, takes $\mathbf{x}^{(3)}(t) = [\mathbf{x}^{(2)}(t),\hat{\mathbf{J}}_{prior}(t),\hat{\mathbf{\Theta}}_{root}(t)]\in \mathbb{R}^{12N_s+3N_j+6}$ as the input and outputs a refined root-relative 3D joint-map with more accurate positions $\hat{\mathbf{J}}_{fine}(t)$, supervised by the loss function
\begin{equation}
    \begin{aligned}
    \mathcal{L}_{fineJ}= \sum_t \parallel \hat{\mathbf{J}}_{fine}(t)-\mathbf{J}^{GT}(t) \parallel_2^2.
    \end{aligned}
\label{equ:loss_fine_joint}
\end{equation}
After that, $\mathbf{x}^{(4)}(t) = [\mathbf{x}^{(3)}(t),\hat{\mathbf{J}}_{fine}(t)]\in \mathbb{R}^{12N_i+3N_j+3N_j+6}$ is fed into \textit{Body-pose Estimator} which learns to solve an IK-like problem to regress joint rotations $\hat{\mathbf{\Theta}}(t)$, supervised by the loss function
\begin{equation}
    \begin{aligned}
    \mathcal{L}_{ik}= \sum_t \parallel\hat{\mathbf{\Theta}}(t)-\mathbf{\Theta}^{GT}(t)\parallel_2^2.
    \end{aligned}
\label{equ:ik_loss}
\end{equation}
Because of the limited expression capability of joint rotations, we introduce the loss \autoref{equ:fk_loss} for \textit{Body-pose Estimator} by Forward Kinematics (FK) to better reconstruct the 3D joint positions.
\begin{equation}
    \begin{aligned}
    \mathcal{L}_{fk}= \sum_t \parallel\operatorname{FK}(\hat{\mathbf{\Theta}}(t)) - \mathbf{J}^{GT}(t)\parallel_2^2.
    \end{aligned}
\label{equ:fk_loss}
\end{equation}
The complete loss function of this module is formulated as
\begin{equation}
    \begin{aligned}
    \mathcal{L}_{pose} = \lambda_3\mathcal{L}_{fineJ} + \lambda_4\mathcal{L}_{ik} + \lambda_5\mathcal{L}_{fk}.
    \end{aligned}
\label{equ:pose_loss}
\end{equation}

\subsection{Pose-guided Translation Correction} 
\label{sec:m2}
% translation estimation is challenging for purely inertial mocap
Global translation estimation, especially in the large-scale scenario, is challenging for purely inertial mocap methods since no localization information can be provided by IMUs. Moreover, IMUs suffer from the drifting problem. Even in the state-of-the-art work (TransPose~\cite{yi2021transpose}), the capture of global movements should be performed under strong assumptions such as level ground and sufficient foot-ground contacts. 
%
% LiDAR is beneficial, but previous work did not make use of it
LiDAR, however, brings significant benefits to this task by measuring distances directly.
LiDARCap~\cite{li2022lidarcap} utilizes the average of points as global translation, however, there exists a deviation between the scanned point cloud and real movement positions, because LiDAR only collects partial points of the performer in the capture view. 
%
% our design
Accurately, the deviation differs for different poses. For example, standing still and bending result in similar global translation positions, while different averages of points. Considering this, we construct the relationships between poses and translation deviations and treat the translation discrepancy as a per-pose variable, which is only correlated with the pose performed. In practice, we combine \textit{encoder} and \textit{estimator} block to model the motion pattern from the point cloud and learn this deviation, rather than regress the global translation directly.
%standing still and bending still give different translation discrepancies, but both of them are nearly constant wherever the performer is in the capture region. 
%Hence, we use $\hat{\mathbf{D}}(t),\ \mathbf{D}^{GT}(t)\in \mathbb{R}^{3}$ to denote the predicted and ground-truth translation discrepancy, respectively. Specifically, $\mathbf{D}^{GT}(t)$ is defined by equation~\ref{equ:offset_def}:
The ground-truth translation discrepancy $  \mathbf{D}^{GT}(t)\in \mathbb{R}^{3} $ can be calculated as follows:
\begin{equation}
    \begin{aligned}
    \mathbf{D}^{GT}(t) = \widetilde{\mathbf{J}}^{GT}_{root}(t) - \operatorname{avg}(\widetilde{\mathbf{x}}^{(1)}(t)),
    \end{aligned}
\label{equ:offset_def}
\end{equation}
where $\widetilde{\mathbf{J}}_{root}^{GT}(t)$ is the ground-truth absolute position of the root joint and the operator $\operatorname{avg}(\cdot)$ calculates the approximate center of given point cloud by averaging the positions of all the points. Finally, we use the loss \autoref{equ:trans_loss} to train this module:
\begin{equation}
    \begin{aligned}
    \mathcal{L}_{trans} = \sum_t \parallel \hat{\mathbf{D}}(t) - \mathbf{D}^{GT}(t) \parallel_2^2,
    \end{aligned}
\label{equ:trans_loss}
\end{equation}
where $\hat{\mathbf{D}}(t)$ denotes the predicted translation discrepancy result.

\begin{figure*}[ht]
	\centering
	\includegraphics[width=\linewidth]{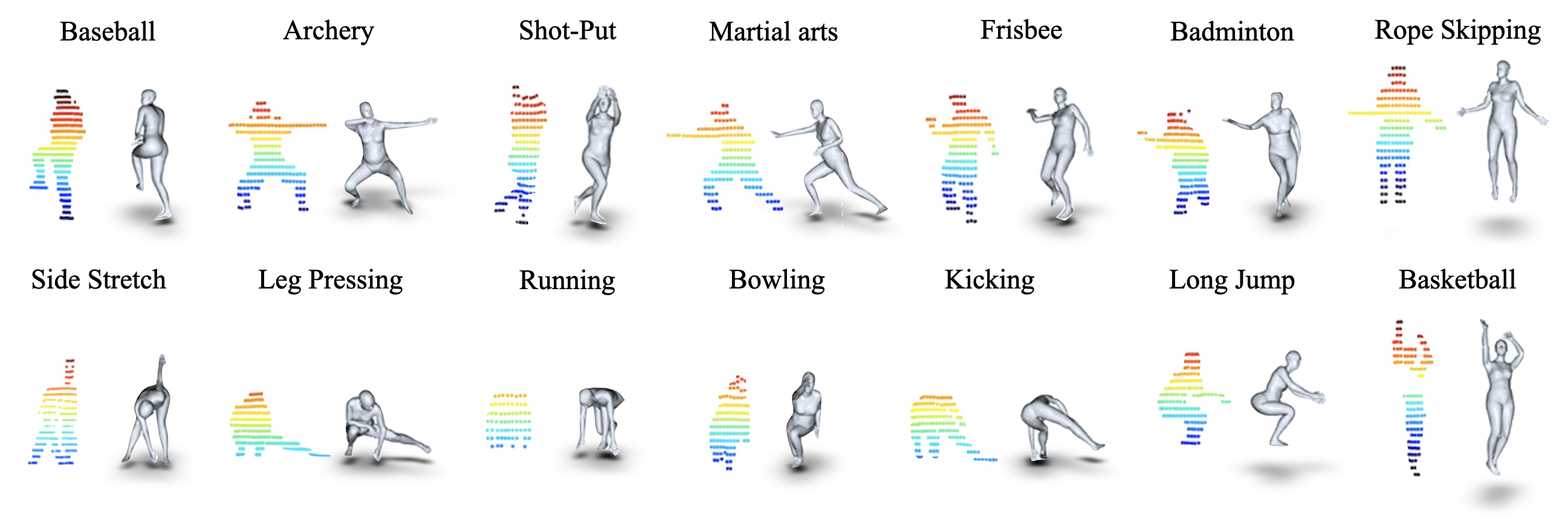}
	\caption{LIPD dataset contains diverse human poses from daily actions to professional sports actions. We demonstrate the point cloud and corresponding ground-truth mesh for several action samples. Such complex poses, like crouching or having the back to LiDAR in the second row, bring challenges for accurate mocap due to severe self-occlusions. We also have multi-person scenarios with external occlusions, as the last basketball case shows, where the player's body is partially covered by others in the LiDAR's view.}
	\label{fig:dataset}
	\vspace{-2ex}
\end{figure*}

\subsection{Implementation Details}
\label{sec:impl_details}
We implement our network using PyTorch 1.8.1 with CUDA 11.1. The training of the two modules is separate, which uses batch size of $32$, sequence length of $32$, learning rate of $10^{-4}$ and decay rate of $10^{-4}$ with AdamW optimizer~\cite{loshchilov2017decoupled}. The weights of loss functions are: $\lambda_1=1,\ \lambda_2=1,\ \lambda_3=0.2,\ \lambda_4=0.7,\ \lambda_5=0.7$. All the training and evaluation are conducted on a server with an Intel(R) Xeon(R) E5-2678 CPU and an NVIDIA RTX3090 graphics card.  Furthermore, we use free acceleration input, which escapes from the influence of gravity in inertial coordinate frame.

\section{Dataset: LIPD}

%篮球 羽毛球 网球 棒球 保龄球 乒乓球 足球 排球 铅球 铁饼 飞盘 跳绳 射箭 武术 太极 跆拳道 拳击 wota art 舞蹈 攀登 跳远 跳高 跨栏 跑步 各类热身动作 演员15人 总共数据量
%62341frame(10fps) train:39593 test:22748 生成数据：746267 LIP(w/o GT):8193

%Our goal is to capture challenging 3D human motions in large-scale scenarios with consistently accurate local pose and global trajectory estimation using single LiDAR and 4 IMUs. Figure~\ref{fig:pipeline} illustrates an overview of the entire pipeline, which consists of two cooperative modules to infer pose and translation, respectively. For pose inference module, we propose a two-stage network working in a coarse-to-fine manner to distill rough human body skeleton and global orientation from raw point cloud first, and then regress human body pose from joint positions refined by IMU measurements (Section~\ref{sec:m1}). For translation inference module, we design a pose-guided approach to learn the latent domain gap between LiDAR measurements and real global movements, through which the inherent translation deviation and random noise from point cloud can be eliminated (Section~\ref{sec:m2}).

%\subsection{Dataset: LIPD}

\begin{table}[ht!]
	\centering
	\caption{Overview of datasets we use. The data mode of point cloud and IMU measurements for each dataset is reported. Moreover, "Capture distance" shows the maximum distance between performer and capture device; "Activity range" stands for the average 3D space size in which the performer moves. Both of them reflect the scale attribute of each dataset.}
	\setlength{\tabcolsep}{1.2mm}
	\begin{tabular}{cccccc}
	    \toprule
    	    Dataset  & \begin{tabular}[c]{@{}l@{}}Point \\cloud \end{tabular}& \begin{tabular}[c]{@{}l@{}}IMU \\recording \end{tabular} & \begin{tabular}[c]{@{}l@{}} Capture \\distance($m$)\end{tabular} & \begin{tabular}[c]{@{}l@{}}Activity \\range($m^3$)\end{tabular}\\
	    \midrule
    	DIP-IMU~\cite{huang2018DIP} & Syn & Real & N/A & N/A\\
    	AMASS~\cite{AMASS_ICCV2019} & Syn & Syn & 3.42 & 0.27\\
    	AIST++~\cite{li2021ai}&  Syn & Syn & 4.23 & 0.67\\
		LiDARHuman26M~\cite{li2022lidarcap} & Real & Syn & 28.05 & 142.95\\ 
		LIPD  & Real & Real & 30.04 & 366.34\\
		\bottomrule
\end{tabular}
\label{tab:dataset}
	\vspace{-2ex}
\end{table}

To learn the local pose and global translation estimator working for large-scale scenarios, a huge LiDAR-IMU hybrid mocap dataset is required, which needs the LiDAR to capture the point cloud data and dense IMU sensors to provide the ground-truth. In this paper, we propose the first long-range LiDAR-IMU hybrid mocap dataset focusing on diverse challenge motions, such as many athletic motions. The dataset contains 15 performers, about 30 kinds of motions and 62,341 frames of LiDAR point cloud and corresponding IMU measurements in total. LIPD also provides ground-truth SMPL pose parameters and RGB images. We divide the data by 39,593 frames as the training set and 22,748 frames as the testing set.

%To learn the local pose and global translation estimator working for large-scale scenarios, a huge LiDAR-IMU hybrid mocap dataset is required, which is unfortunately unavailable at present to the best of our knowledge. Due to the inconvenience and constraint on performers to act in arbitrary motions, it is difficult to collect large amounts of data with dense IMU sensors. Thus, we simulate plenty of synthetic LiDAR-IMU measurements with ground-truth SMPL pose parameters on DIP-IMU, LiDARHuman26M, AIST++ and a subset of AMASS, including ACCAD, BMLMovi, CMU, and TotalCapture. For simulation details and rationality analysis, please refer to the appendix.

% instruction and modification %
%\yuexin{Why does simulation can solve the problem? Any try to reduce the gap to real data, such as the rationality of the simulation of sensors, the different ranges, and the diversity of poses? Give a detailed explanation to eliminate the confusion in your training set.}
%\zhaochf{For simulation details, please refer to the appendix.}
%%%%%%%%%%%%%%%%%%%%%%%%%%%%%
We collect LIPD by Noitom IMU sensors~\cite{noitom} and OUSTER-1 LiDAR~\cite{OUSTER}, performers wear a full set of Notiom equipment to collect themed challenge actions within the range of 12-30$m$. Furthermore, to show the generalization capability of our method, we record extra 8,193 frames of data by our proposed light-weight setting, including single LiDAR and four Xsens Dot IMU sensors\cite{XSENS}, in diverse wild scenes for visualization evaluation. Xsens provides professional applications that can record offline IMU data on mobile phones, which means that actors can move freely without space restrictions. Performers only need to wear four Xsens Dot IMU sensors in four limbs for testing, so that they can perform more challenging motions.

\subsection{Characteristic.}
\textbf{Large-scale scenes.}
As mentioned in \autoref{tab:dataset}, LIPD has 366.34$m^3$ activity range, not only providing the long-range data, but also covering scenes in varying heights, such as the climbing motion on stairs. Meanwhile, LIPD provides more views for motion capture compared with LiDARHuman26M~\cite{li2022lidarcap}, which only includes the top view. Furthermore, we also provide the scene in dark to show that the LiDAR can capture accurate point cloud data in extreme environments, which is superior to cameras.
\\
\textbf{Diverse motions. } 
The LiDAR-only mocap dataset, LiDARHuman26M~\cite{li2022lidarcap}, contains about 20 daily human motions without too many occlusions. Due to the view-dependent property of LiDAR, the performance drops dramatically when handling complex motions with obvious occlusions. LIPD is a new benchmark for mocap in large-scale scenes involving diverse challenging poses, as \autoref{fig:dataset} shows. It includes both raw LiDAR point cloud and IMU measurements for facilitating the research of accurate mocap in large-scale scenes based on multimodal inputs.
%At present, the existing LiDAR-IMU hybrid dataset only contains a variety of daily motions. To avoid more self-occlusion, LiDARHuman26M uses the form of a top view. LIPD provides more than twenty motion actions with a large number of self-occlusion cases, which leads to more sparse point cloud data and also means that actions are more difficult to predict. At the same time, we also select complex dance motions from existing mocap datasets, such as ACCAD and AIST++ to enrich LIPD.

%Large scale scenes result in extremely sparse point clouds, meanwhile it includes point clouds from different perspectives. LIPD even includes the extreme situation of less than 10 points. In addition, we focus on challenging motions. These motions produce a lot of self-occlusion, leading to missing of human body, which seriously interferes with our motion prediction. Different motions may result in similar point cloud data due to the absence of point cloud at a certain part.

\subsection{Extension. }
In order to enrich the dataset with more different motions, we simulate plenty of synthetic LiDAR-IMU measurements with ground-truth SMPL~\cite{SMPL2015} pose parameters on DIP-IMU~\cite{huang2018DIP}, LiDARHuman26M~\cite{li2022lidarcap}, AIST++~\cite{li2021ai} and a subset of AMASS~\cite{AMASS_ICCV2019}, including ACCAD, BMLMovi, CMU, and TotalCapture~\cite{trumble2017total}. The synthesized data consists of 74,6267 frames with 4,238 motion genres. As for the protocol of dataset splitting, we take DIP-IMU~\cite{huang2018DIP}, TotalCapture~\cite{trumble2017total}, the testing set of LiDARHuman26M~\cite{li2022lidarcap}, and the testing set of LIPD for evaluation. The left data is for training. \autoref{tab:dataset} gives an overview of all the datasets. For simulation details and rationality analysis, please refer to the appendix.

\subsection{Challenge. }
Based on LiDAR and sparse IMUs, LIPD proposes a novel light-weight sensor setting for mocap in large-scale scenes. However, there are two main challenges, need to be solved, for reconstructing accurate human motions. The first is extracting valuable features from a sparsity-varying LiDAR point cloud, where the person in more than 20m away only contains a few points, and the second is finding an effective sensor-fusion method to make LiDAR and IMUs actually complement each other. We explore potential solutions for these problems and propose LIP as the baseline for this task.
\section{Experiments}
\label{sec:experiments}

\begin{table*}[ht!]\scriptsize
	\centering
	\caption{Quantitative comparisons between LIP and related methods on our evaluation dataset. Note that LiDARHuman26M~\cite{li2022lidarcap} and LIPD dataset only contains data with rate of 10fps, which is not applicable(N/A) for TransPose~\cite{yi2021transpose}. Also, the CD metric is not used for DIP-IMU~\cite{huang2018DIP} dataset since no translation is provided.}
	\setlength{\tabcolsep}{1.2mm}
	\begin{tabular}{cccccccccccccccc}
	    \toprule
	    \multirow{2}{*}{} & \multicolumn{4}{c}{TotalCapture~\cite{trumble2017total}} &  \multicolumn{3}{c}{DIP-IMU~\cite{huang2018DIP}} & \multicolumn{4}{c}{LiDARHuman26M~\cite{li2022lidarcap}}&
	    \multicolumn{4}{c}{LIPD(Ours)}\\
	    \cmidrule(r){2-5}
	    \cmidrule(r){6-8}
	    \cmidrule(r){9-12}
	    \cmidrule(r){13-16}
	    & MPJPE & Mesh Err & Ang Err & CD & MPJPE & Mesh Err & Ang Err & MPJPE & Mesh Err & Ang Err & CD & MPJPE & Mesh Err & Ang Err &CD\\
	    \midrule
	   % ICP & 689.6 & 767.1 &45.1  & 0.8 & 712.9 &791.8 &41.8 &N/A &648.9 &727.6 & 63.0&1.0\\
		LiDARCap\cite{li2022lidarcap} & 45.0 & 56.6 &8.5 & 6.4 & 41.5 & 54.9 & 12.4 & 78.7 & 94.8 & 21.0 & 8.1 &69.4&85.5&12.5&7.9  \\
		TransPose\cite{yi2021transpose} & 55.7 & 63.6 & 12.3 & 102.7 & 49.0 & 58.3 & 7.6 & N/A & N/A & N/A & N/A & 76.8 & 87.1 & 17.3  & 736.9\\
		\textbf{LIP} & \textbf{30.0} & \textbf{39.5} & \textbf{7.4} &\textbf{0.7}& \textbf{30.2}&\textbf{39.1}&\textbf{9.6}& \textbf{60.7}&\textbf{71.6}&\textbf{11.6}&\textbf{3.5}&
		\textbf{48.9}&\textbf{59.8}&\textbf{11.3}&\textbf{3.2}\\
		\bottomrule
\end{tabular}
\label{tab:Compare}
\end{table*}

\begin{figure*}[ht!]
    \centering
    \includegraphics[width=\linewidth]{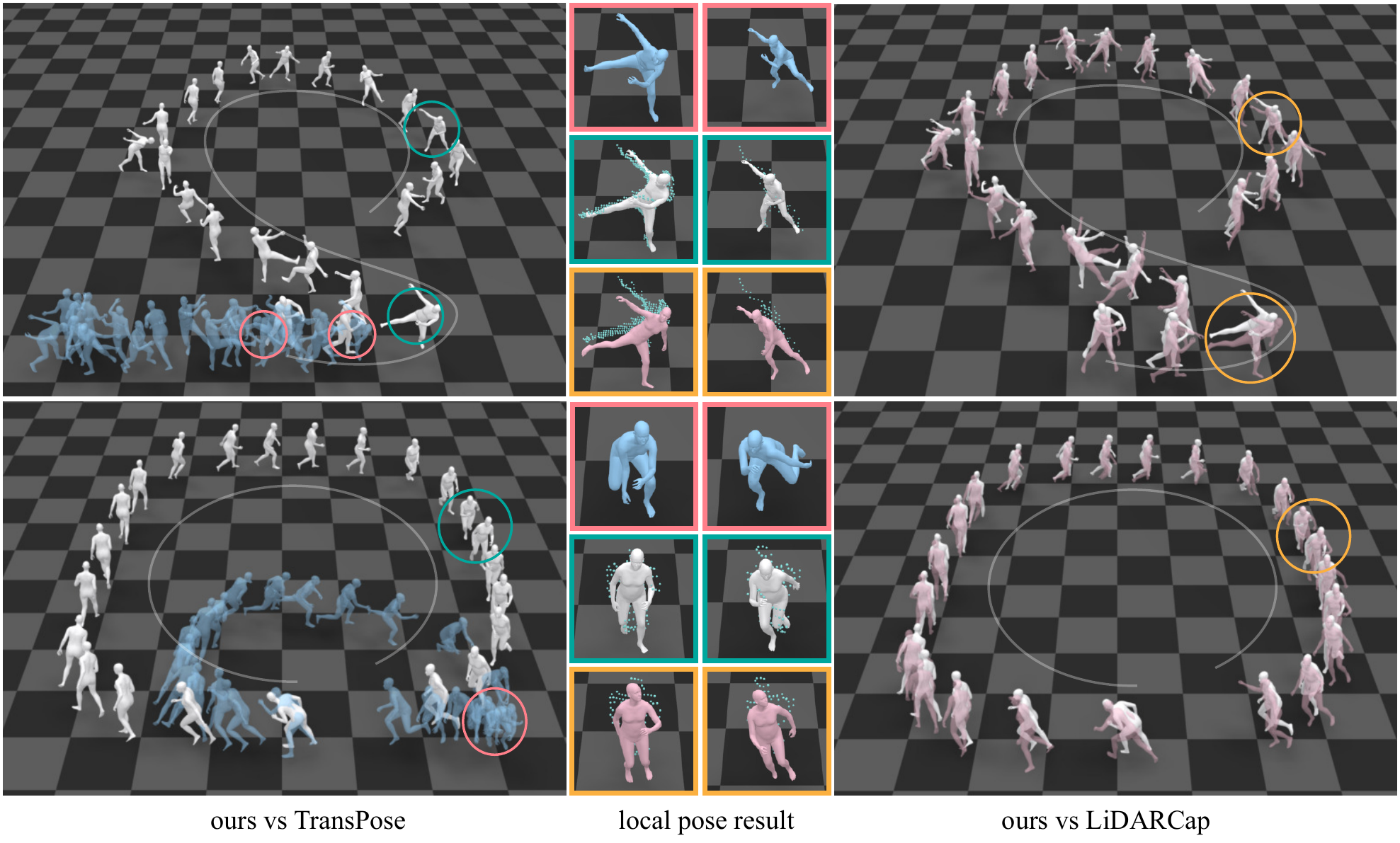}
    \caption{Qualitative results of LIP (white) compared with TransPose~\cite{yi2021transpose} (blue) and LiDARCap~\cite{li2022lidarcap} (pink). The trajectory is sketched by a grey curve. Detailed comparisons of some key frames are circled and zoomed in to show the local pose and alignment with point cloud. For intuitive visualization, we use the arithmetic mean of raw point cloud as the global translation for LiDARCap~\cite{li2022lidarcap} in which translation inference is not included.}
    \label{fig:comp}
    \vspace{-2ex}
\end{figure*}

\begin{figure*}[ht!]
    \centering
    \includegraphics[width=\linewidth]{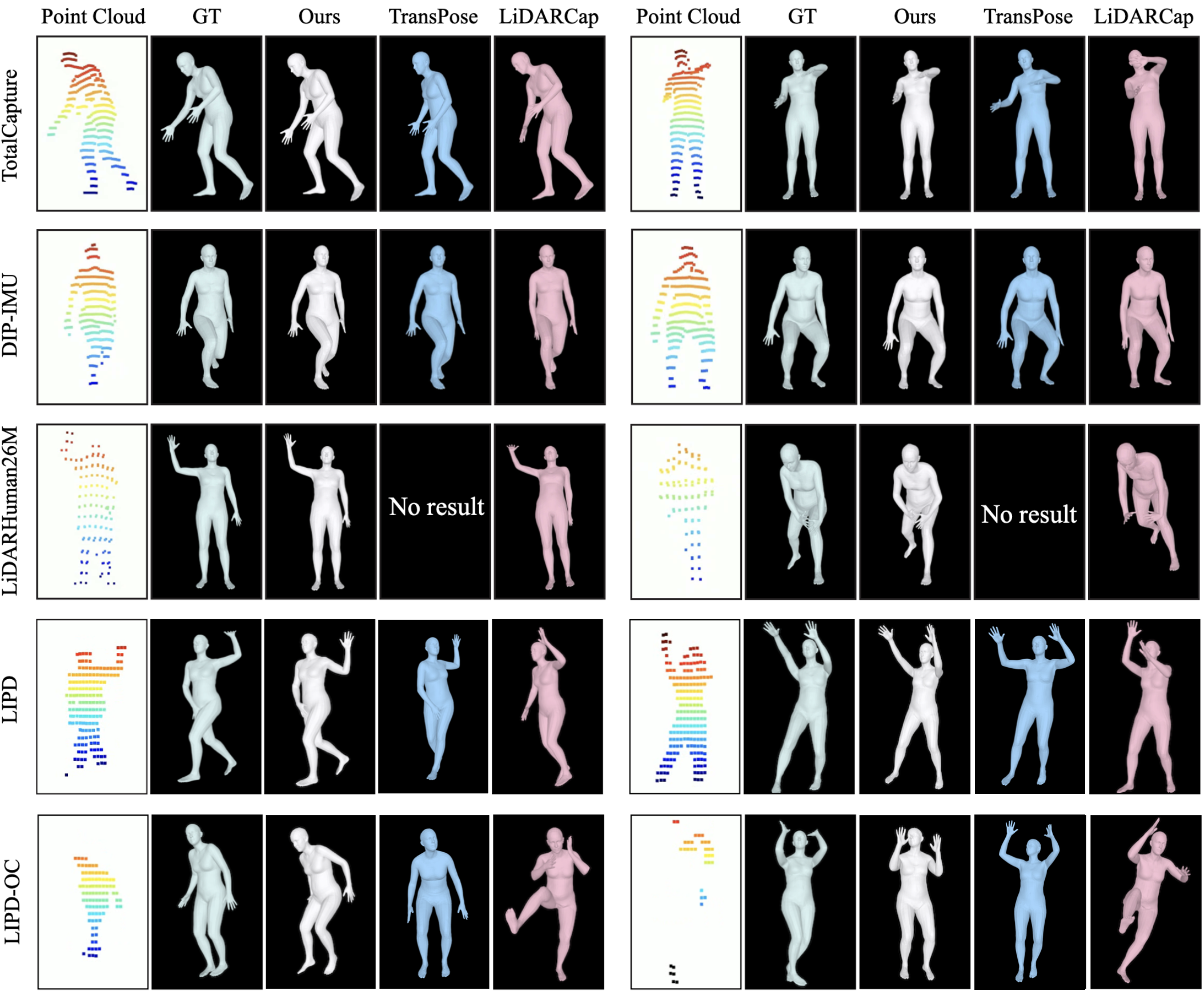}
    \caption{Qualitative comparison of capture performance for local details. Our method outperforms state-of-the-art works by more accurate local pose estimation on three different synthetic datasets and LIPD. LIPD-OC means the data with severe occlusions in LIPD. Note that LiDARHuman26M~\cite{li2022lidarcap} is not applicable for TransPose~\cite{yi2021transpose} since it only provides 10fps mocap data, while frame rate of 60fps is required by TransPose~\cite{yi2021transpose}.}
    \label{fig:local_comparison}
    \vspace{-3ex}
\end{figure*}

\begin{figure*}[t]
    \centering
    \includegraphics[width=\linewidth]{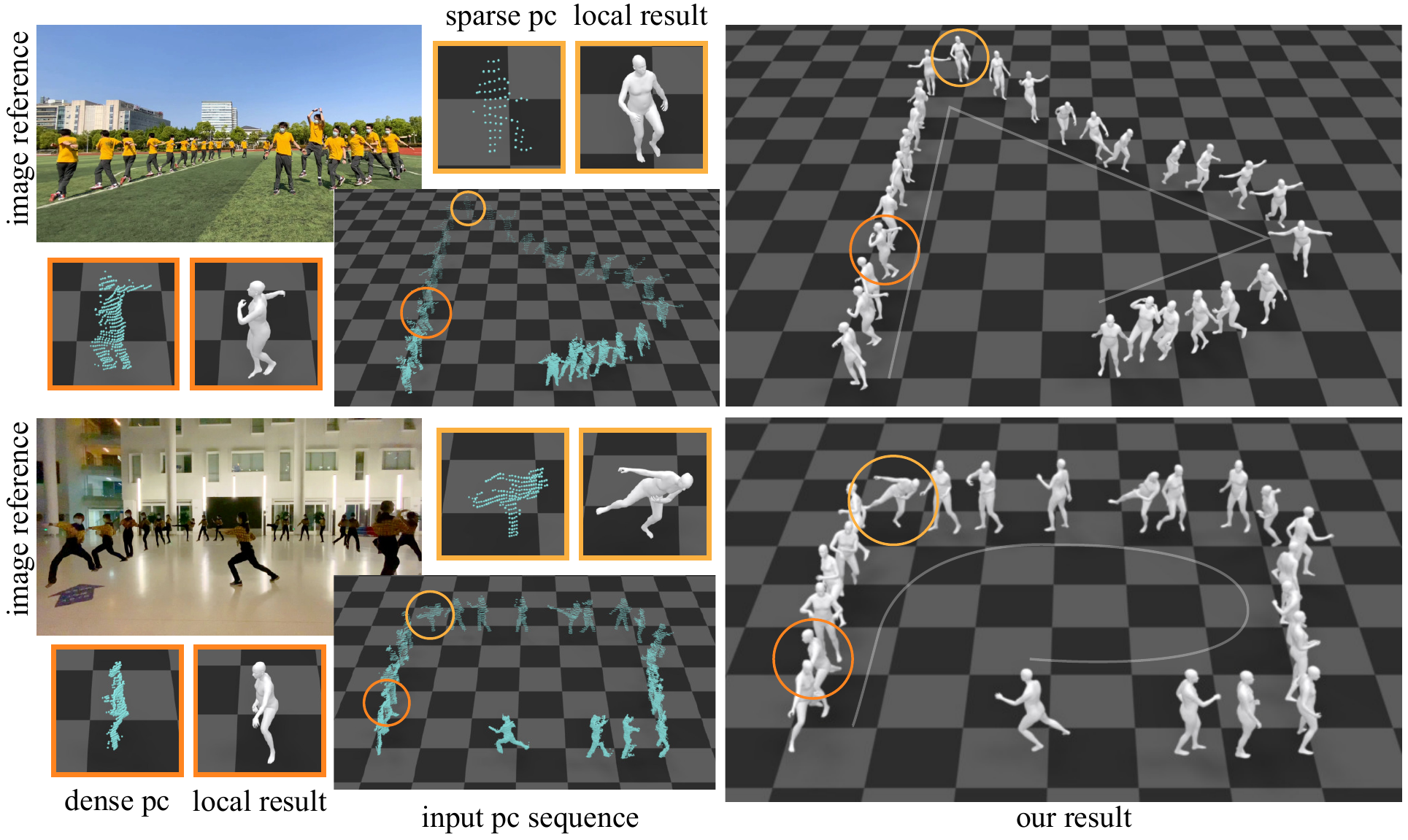}
    \caption{Our results. We provide image reference, point cloud (pc) input, and mocap results in 3D scene in this figure. Some key frames are circled and zoomed in to show detailed local poses.}
    \label{fig:gallery}

\end{figure*}

\begin{table*}[ht!]\scriptsize
 	\centering
 	\caption{Ablation study of different fusion strategies and the \textit{Joint-map Estimator} (JE) block in our network.}
 	\setlength{\tabcolsep}{1.75mm}
 	\begin{tabular}{ccccccccccccc}
 	    \toprule
 	    \multirow{2}{*}{Local pose} & \multicolumn{3}{c}{TotalCapture~\cite{trumble2017total}} &  \multicolumn{3}{c}{DIP-IMU~\cite{huang2018DIP}} & \multicolumn{3}{c}{LiDARHuman26M~\cite{li2022lidarcap}}&
 	    \multicolumn{3}{c}{LIPD} \\
 	    \cmidrule(r){2-4}
 	    \cmidrule(r){5-7}
 	    \cmidrule(r){8-10}
 	    \cmidrule(r){11-13}
 	    & MPJPE & Mesh Err & Ang Err & MPJPE & Mesh Err & Ang Err & MPJPE & Mesh Err & Ang Err &MPJPE & Mesh Err & Ang Err\\
 	    \midrule
 	    FM-1 &43.2 &55.1 &12.9 &43.6 &54.9 & 15.4&\textbf{59.4} &72.7 &17.3 &49.5 &61.4 &13.2\\
 	    FM-2 &32.3 &42.3 &7.8 & 31.2&41.3 &9.7 &61.8 &74.3 &14.2 &50.9 &62.3 &11.3\\
 	    FM-3 & 47.8&62.3&8.1&44.6 &58.3 &11.4 &67.6 &83.4 &14.2&54.9&67.3&11.5\\
 	    FM-4 & 36.7& 42.4& \textbf{6.7}& 37.3&49.0 &10.0 &59.6 &\textbf{70.8} &12.6 & 52.4&64.5 &\textbf{11.0}\\
 		\textbf{Ours}&\textbf{30.0}&\textbf{39.5} &7.4 &\textbf{30.2}&\textbf{39.1}&\textbf{9.6}&60.7 &71.6 &\textbf{11.6}&\textbf{48.9}&\textbf{59.8}&11.3 \\
 		Ours w/o JE & 35.5&45.5&8.6& 34.3&43.3&10.2&61.2&72.5&11.6&51.4&62.8&11.3\\
 		\bottomrule
 		
\end{tabular}
\label{tab:ablation}
 \vspace{-3ex}
\end{table*}

\begin{table}[ht!]\scriptsize
 	\centering
 	\caption{Ablation study of different global translation methods and our pose information guided processing (PG) in LIP. We demonstrate the evaluation results using CD metric.}
  %is evaluated together with average translation estimation and deirect regression, using only CD metric.}
 	\setlength{\tabcolsep}{1.5mm}
 	\begin{tabular}{ccccccccccccc}
 	    
 		Global translation & \multicolumn{3}{c}{TotalCapture~\cite{trumble2017total}} & \multicolumn{3}{c}{LiDARHuman26M~\cite{li2022lidarcap}} & \multicolumn{3}{c}{LIPD}\\
 		\midrule
 		Average estimation & \multicolumn{3}{c}{6.4} & \multicolumn{3}{c}{8.1} & \multicolumn{3}{c}{7.9}\\
 		ICP estimation & \multicolumn{3}{c}{2.0} & \multicolumn{3}{c}{4.7} & \multicolumn{3}{c}{4.1}\\
 		Ours w/o PG & \multicolumn{3}{c}{0.9} & \multicolumn{3}{c}{3.7} & \multicolumn{3}{c}{3.4}\\
 		\textbf{Ours} & \multicolumn{3}{c}{\textbf{0.7}} & \multicolumn{3}{c}{\textbf{3.5}} & \multicolumn{3}{c}{\textbf{3.2}}\\
 		\bottomrule
\end{tabular}
\label{tab:trans_ablation}
\vspace{-2ex}
\end{table}

In this section, we compare our method with State-Of-The-Art (SOTA) methods qualitatively and quantitatively to show the superiority of the proposed LIP in \autoref{sec:comparison}. 
In addition, in \autoref{sec:evaluation}, sufficient ablation studies are conducted to further demonstrate the rationality of the strategy for fusing multi-modal data and our architecture design. Moreover, we test multiple input configurations to prove the single LiDAR with four IMU sensors is an appropriate setting. For evaluation metrics, we use 1) \textbf{MPJPE($mm$)}: mean per root-relative joint position error; 2) \textbf{Mesh Err($mm$)}: mean per SMPL mesh vertex position error; 3) \textbf{Ang Err($deg$)}: mean per global joint rotation error to evaluate local pose, and 4) \textbf{CD($cm$)}: chamfer distance between vertices of SMPL mesh result and raw point cloud to evaluate global translation. For all these metrics, \textbf{lower means better}.

\subsection{Comparison}
\label{sec:comparison}

\begin{figure*}[ht!]
\centering
\includegraphics[width=\linewidth]{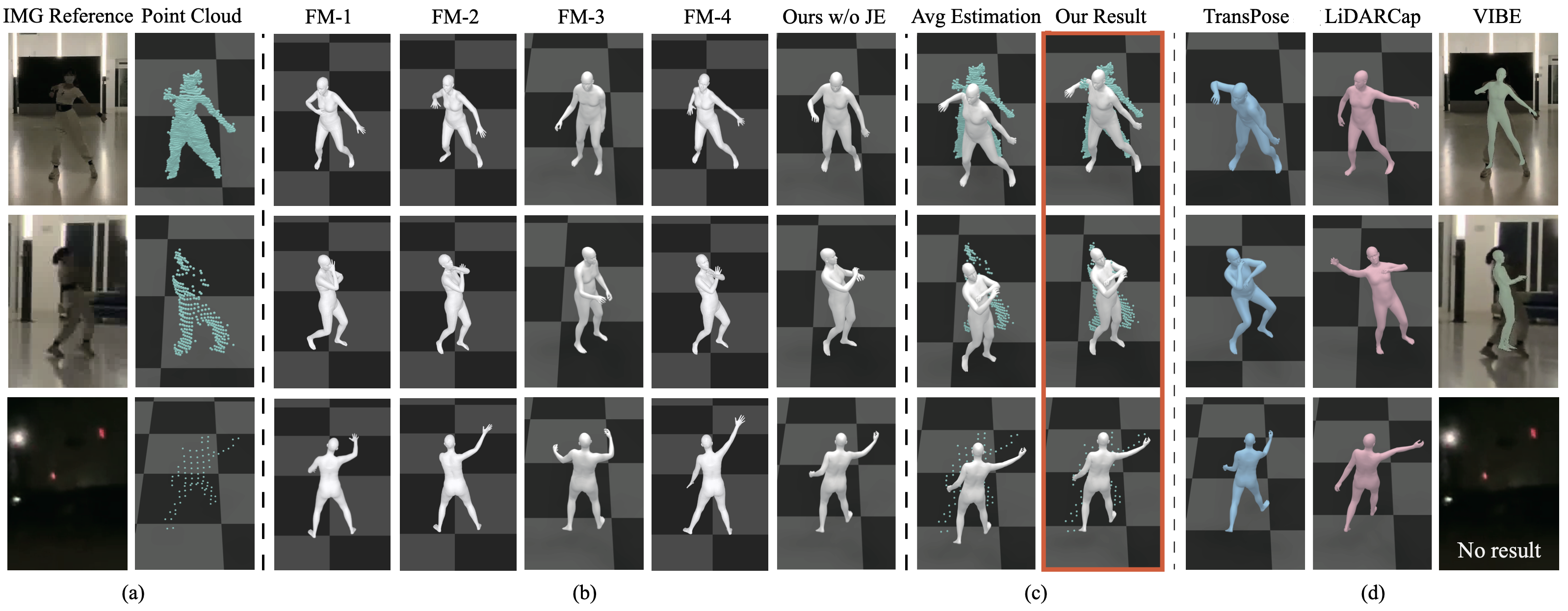}
\caption{Qualitative evaluations of our network design. We provide (a) image reference and point cloud input; (b) local pose evaluation; (c) global translation evaluation; (d) results of other methods.}
\vspace{-2ex}
\label{fig:eval}
\end{figure*}

\begin{table*}[ht!]\scriptsize
 	\centering
 	\caption{Ablation study for input configuration. We evaluate our choice of single LiDAR with 4 IMUs by experimenting the local motion estimation accuracy of various combinations of input modalities.}
 	\setlength{\tabcolsep}{1.7mm}
 	\begin{tabular}{ccccccccccccc}
 	    \toprule
 	    \multirow{2}{*}{} & \multicolumn{3}{c}{TotalCapture~\cite{trumble2017total}} &  \multicolumn{3}{c}{DIP-IMU~\cite{huang2018DIP}} & \multicolumn{3}{c}{Lidarhuman26M~\cite{li2022lidarcap}}& \multicolumn{3}{c}{LIPD} \\
 	    \cmidrule(r){2-4}
 	    \cmidrule(r){5-7}
 	    \cmidrule(r){8-10}
 	    \cmidrule(r){11-13}
 	    & MPJPE & Mesh Err & Ang Err  & MPJPE & Mesh Err & Ang Err  & MPJPE  & Mesh Err & Ang Err &MPJPE  & Mesh Err & Ang Err \\
 	    \midrule
 	    LiDAR Only &35.5&47.9&12.33 &35.1&47.0&15.41 &68.8&87.1&20.20&60.1&76.1&14.49 \\
         4 IMUs Only &76.3&93.2&10.65&91.4&103.1&14.04&90.0&106.6&14.14&69.1&83.8&12.77\\
 		5 IMUs only & 59.9&74.0&9.59&79.0&88.8&11.96 &71.9&87.4&12.86&50.3&60.7&11.65\\
 		LiDAR+2 IMUs & 31.1&41.0 &9.38 &31.1&41.1&11.44 & 63.1&76.8&14.53 &53.9&67.1&12.64\\
 		\textbf{LiDAR+4 IMUs(Ours)} &30.0&39.5 &7.4 &30.2&39.1&9.6&60.7 &71.6 &11.6&48.9&59.8&11.3\\
 		LiDAR+5 IMUs &30.1&39.7&7.4&30.7&39.9&9.7&57.9&69.0&11.2&44.9&55.4&10.6 \\
 		LiDAR+12 IMUs & 23.6&32.8&3.63 &20.8 &28.4 &5.76 & 49.0&59.3 &7.45&25.1&32.1&8.4 \\
 		\bottomrule
\end{tabular}
\label{tab:input}
%\vspace{-2ex}
\end{table*}

%\iffalse
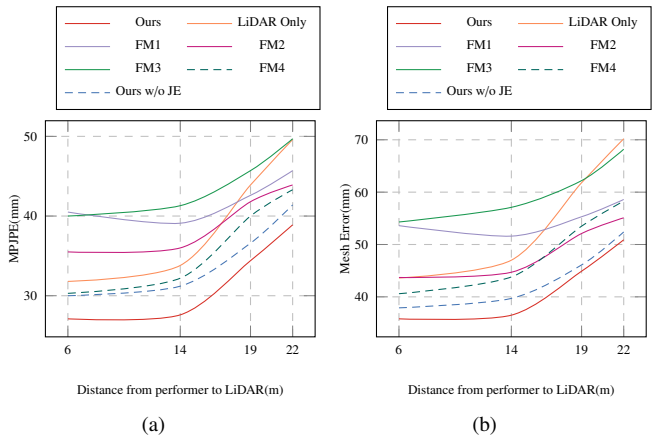
\begin{figure}[ht!]
    \centering
    \subfigure[]{
    \begin{tikzpicture}
    \pgfplotsset{every axis legend/.append style={
    at={(1.02,1.53)}},every axis y label/.append style={at={(0.23,0.5)}}}
    \begin{axis}[xlabel=Distance from performer to LiDAR(m),ymajorgrids=true,xmajorgrids=true,
    grid style=dashed,ylabel=MPJPE(mm),xtick=data,legend columns=2,legend style={font=\tiny},font=\tiny,width=.585\linewidth]
    \addplot+[smooth][color=c1,mark=circle] table [x=dis,y=joint,col sep=comma]{mydata.csv};
    \addplot+[smooth][color=c2,mark=circle] table [x=dis,y=Pl_j,col sep=comma]{mydata.csv};
    \addplot+[smooth][color=c3,mark=circle] table [x=dis,y=FM1_j,col sep=comma]{mydata.csv};
    \addplot+[smooth][color=c4,mark=circle] table [x=dis,y=FM2_j,col sep=comma]{mydata.csv};
    \addplot+[smooth][color=c5,mark=circle] table [x=dis,y=FM3_j,col sep=comma]{mydata.csv};
    \addplot+[smooth][color=c6,mark=circle] table [x=dis,y=FM4_j,col sep=comma]{mydata.csv};
    \addplot+[smooth][color=c7,mark=circle] table [x=dis,y=wo_j,col sep=comma]{mydata.csv};
    \legend{Ours,LiDAR Only,FM1,FM2,FM3,FM4,Ours w/o JE}
    \end{axis}
    \end{tikzpicture}}
    \subfigure[]{
    \begin{tikzpicture}
    \pgfplotsset{every axis legend/.append style={
    at={(1.02,1.53)}},every axis y label/.append style={at={(0.23,0.5)}}}
    \begin{axis}[xlabel=Distance from performer to LiDAR(m),ymajorgrids=true,xmajorgrids=true,
    grid style=dashed,ylabel=Mesh Error(mm),xtick=data,legend columns=2,legend style={font=\tiny},font=\tiny,width=.585\linewidth]
    \addplot+[smooth][color=c1,mark=circle] table [x=dis,y=mesh,col sep=comma]{mydata.csv};
    \addplot+[smooth][color=c2,mark=circle] table [x=dis,y=Pl_m,col sep=comma]{mydata.csv};
    \addplot+[smooth][color=c3,mark=circle] table [x=dis,y=FM1_m,col sep=comma]{mydata.csv};
    \addplot+[smooth][color=c4,mark=circle] table [x=dis,y=FM2_m,col sep=comma]{mydata.csv};
    \addplot+[smooth][color=c5,mark=circle] table [x=dis,y=FM3_m,col sep=comma]{mydata.csv};
    \addplot+[smooth][color=c6,mark=circle] table [x=dis,y=FM4_m,col sep=comma]{mydata.csv};
    \addplot+[smooth][color=c7,mark=circle] table [x=dis,y=wo_m,col sep=comma]{mydata.csv};
    \legend{Ours,LiDAR Only,FM1,FM2,FM3,FM4,Ours w/o JE}
    \end{axis}
    \end{tikzpicture}}
    \caption{Evaluation for the generalization capability on various distances on MPJPE and Mesh Err.}
\label{fig:generalization}
\vspace{-2ex}
\end{figure}
%\fi

We conduct comparisons to illustrate that our proposed LIP method enables more accurate capture for challenging motions and more precise translations in large-scale spaces. 
%For further analysis, quantitative and qualitative comparisons with other IMU-based methods and LiDAR-based methods are conducted.
We select current SOTA TransPose~\cite{yi2021transpose} and LiDARCap~\cite{li2022lidarcap} for comparative analysis. We use the model provided by TransPose~\cite{yi2021transpose} and reproduced LiDARCap~\cite{li2022lidarcap} network (no released model) for comparison. Note that we test our reproduced model on the setting of LiDARCap~\cite{li2022lidarcap} and achieves comparable performance as its paper claims. Thus, the comparison is fair. The superior results of LIP with different evaluation metrics are illustrated in \autoref{tab:Compare}.
%Tab.\ref{tab:Compare} illustrates the results between our method and others with different evaluation metrics. Since we combine the LiDAR and sparse sensors as the hybrid modal input, we select TransPose and LiDARCap for comparative analysis, which proves that our network structure makes effective use of the advantages of multi-modality and achieves significant superiority over state-of-the-arts.
%We report the Mean Per Joint Position Error(MPJPE), mesh error, angular error, and jitter error for evaluating the accuracy of motion capture.
The visualization evaluation for mocap large-scale scenes is shown in \autoref{fig:comp}. Benefiting from LiDAR-IMU hybrid modal input and our coarse-to-fine design, our method outperforms others by an obvious margin. 
% Add more details and descriptions after the figures are ok.
To take advantage of the 3D distinguish-ability of the LiDAR point clouds, we design the \textit{Pose-guided Translation Correction}. From the global view, LIP provides more accurate localization and more nature motion sequences. For the pose reconstruction of each frame, our result is aligned well with the point cloud. 
%We use our LIPD dataset which consists of abundant challenge motions with large-scale spaces to compare the estimated translation. Our method performs much better than TransPose and LiDARCap as shown in Figure~\ref{fig:comp}.
\autoref{fig:local_comparison} further demonstrate the visualization results of local pose estimation on the testing data for more detailed comparison. Our method is obviously superior to others with results more close to the ground truth, especially for the challenging motions in LIPD. It is worth noting that LIP is still robust for some extreme cases with severe self-occlusions and external occlusions, like cases in the last row of \autoref{fig:local_comparison} shows. In addition, we show more fancy results of our method for mocap in large-scale scenes in \autoref{fig:gallery} to verify the generalization capability of LIP.

\subsection{Ablation Study}

We validate the effectiveness of our sensor-fusion method, network design, and the reasonability of the configuration of our multi-modal hybrid input by ablation studies. 
%Qualitative evaluations are illustrated in Figure~\ref{fig:eval}.
%We validate the effectiveness and generalization ability by apply two ablation experiments. The first experiment verify the effectiveness of the designed coarse-to-fine network structure, and the second experiment validate the necessity of the integration of multi-modal hybrid input and the reasonableness of the input IMU number in our setting.
\label{sec:evaluation}

%\begin{table}[ht]\scriptsize
%\vspace{-2ex}
%    \centering
%    \caption{Ablation study on network architecture. "RE" means \textit{Global pose-prior Regressor} and "JE" indicates \textit{Joint-map Estimator}.}
%    \setlength{\tabcolsep}{2.8mm}
%    \begin{tabular}{cccc}
%    \toprule
%    Pose estimation & MPJPE & Mesh Err & Ang Err\\
%    \midrule
%    Ours w/o RE & 58.6 & 70.7 & 11.5\\
%    Ours w/o JE & 38.1 & 47.7 & 9.9\\
%    Ours & \textbf{31.2} & \textbf{40.0} & \textbf{9.0}\\
%    \bottomrule
%    \toprule
%    \multicolumn{2}{c}{Translation estimation} & & \multicolumn{1}{c}{CD}\\
%    \midrule
%    \multicolumn{2}{c}{Average estimation} & & \multicolumn{1}{c}{7.2}\\
%    \multicolumn{2}{c}{Ours w/o PG} & & \multicolumn{1}{c}{2.1}\\
%    \multicolumn{2}{c}{Ours} & & \multicolumn{1}{c}{\textbf{1.9}}\\
%    \bottomrule
%    \end{tabular}
%\label{tab:ablation}
%\end{table}

\textbf{Ablation Study on Network Designs.} 
%In the network design, we skillfully use different modal inputs in two stages of the network, so that 
Based on the LiDAR-IMU multi-sensor configuration, it is important to make full use the information captured by both sensors and design an effective fusion method to make them complement each other.
Thus, we first demonstrate the superiority of our LiDAR-IMU sensor-fusion method and compare with four other fusion strategies: \textbf{FM-1}) In the data processing phase, we directly merge the point cloud data and IMU data, which is a data-level fusion strategy; \textbf{FM-2}) We use GRU to extract high-dimensional features of IMU data and integrate it with the skeleton joints and root orientation estimated in the first stage from point cloud data; \textbf{FM-3}) The temporal features obtained from the raw point cloud are directly combined with IMU data as the input of the second stage; \textbf{FM-4}) We extracts features of IMU data and point cloud data respectively, and combine different modal features in high-dimensional feature space, and then regress the pose parameters.
%We demonstrate the superiority of our two-stage coarse-to-fine structure and evaluate the following two variants of our network: 1) the temporal features obtained from the raw point cloud are directly combined with IMU data as the input of the second stage, without the coarse joints and root orientation estimation; 2) the model directly regresses the pose parameters from coarse joint estimation in stage one, without any refinement by IMUs. 
We compare the above experiments on the testing set mentioned in \autoref{tab:ablation}. It illustrates that direct data-level or feature-level fusion strategies do not work well due to the large domain gap between these two modal data. Our coarse-to-fine fusion strategy achieves SOTA performance by regressing the coarse skeleton joints and root orientation as the intermediate bridge to align two modalities.
%It can be observed that point cloud data and IMU data cannot be combined directly due to the large discrepancy of the multi-modal data, meanwhile the high-dimensional features may cause mutual interference, so that the dimension reduction task is necessary. We regress the coarse skeleton joints and root orientation as the intermediate task helps the pose estimation, which is more intuitive to guide the subsequent generation of finer joint positions instead of using high-dimensional features. 

The second refinement stage of our network further benefits more accurate results in the estimation of joints and vertices. To verify the effectiveness of \textit{Pose-guided Translation Correction} module, we conduct comparison with other location estimation methods and ablation studies. As \autoref{tab:trans_ablation} shows, the design of deviation regression and the usage of pose features benefit the performance a lot.
To evaluate the generalization capability of LIP to point clouds captured in various ranges with different sparsity, we simulate a bunch of point clouds from TotalCapture~\cite{trumble2017total} dataset at multiple virtual capture distances from 6 to 25 meters. \autoref{fig:generalization} illustrates that our LIP performs consistently well in a wide range of about 20 meters. %Meanwhile, various curves in %Figure~\ref{fig:generalization} further demonstrate the generalization capability of our method.
%Another observation from Table~\ref{tab: ablation} is that the coarse-to-fine design greatly helps the task, which leads to more accurate results in the estimation of joints and vertices.
%\begin{table}
%    \centering
%    \renewcommand{\arraystretch}{1.2}
%    \vspace{6.2ex}
%    \caption{Ablation study on input configuration.}
%    \setlength{\tabcolsep}{1.8mm}
%    \begin{tabular}{cccc}
%    \toprule
%    & MPJPE & Mesh Err & Ang Err\\
%    \midrule
%    LiDAR Only & 32.9 & 43.9 & 14.7\\
%    4 IMUs Only & 87.4 & 98.6 & 13.2\\
%    5 IMUs Only & 74.5 & 85.2 & 11.4\\
%    LiDAR+2 IMUs & 31.8 & 41.6 & 10.8\\
%    LiDAR+4 IMUs(Ours) & \textbf{31.2} & \textbf{40.0} & \textbf{9.0}\\
%    LiDAR+5 IMUs & 30.3 & 39.3 & 8.8\\
%    LiDAR+12 IMUs & 30.2 & 39.0 & 5.5\\
%    \bottomrule
%    \end{tabular}
%\label{tab:input}
%\vspace{-2ex}
%\end{table}

\textbf{Ablation Study on Input Configuration.} 
% overview
To verify that single LiDAR with 4 IMUs is a necessary and compact configuration for high-quality mocap, we experiment various combinations of this two modal inputs with the same network architecture and training strategy. \autoref{tab:input} gives quantitative comparisons which demonstrate that LiDAR-IMU hybrid input overall outperforms single modality input. 
%
% why LiDAR-IMU hybrid input is necessary
Specifically, direct measurements of 3D scene provided by LiDAR reduce the ambiguities of regression from sparsely local inertia to 3D joint positions, which so that decrease the MPJPE and mesh error by more than 30 and 40 millimeters and angular error by 2 degrees over purely inertial input in average. Meanwhile, with the aid of local inertia, inaccurate estimation of joint positions and rotations on point cloud can be corrected in detail. Especially for LIPD datasets, of which the point cloud data is real and imperfect with random noise, LIP improves the performance of LiDAR-only input by about 10 millimeters lower MPJPE, 15 millimeters lower mesh error and 3 degrees lower angular error. In addition, it is necessary to specify that the LiDAR-only performance comes from the same pipeline as LiDAR+$n$ IMUs, except that no IMU data is fused with point cloud information.Different from LiDARCap~\cite{li2022lidarcap}, which is a one-stage method, the Module-I \textit{Multi-modal Pose Estimation} in our pipeline estimates local body motion in two-stage.
%
% why single LiDAR with 4 IMUs is our choice 
Besides, \autoref{tab:input} illustrates that 4-IMU-aid configuration brings considerable improvements with only two more sensors compared with 2-IMU-aid version, and keeps acceptable performance gap to even 12-IMU-aid setting. Therefore, 4 IMU sensors is an appropriate choice to provide sufficient inertial aid while maintaining a light-weight capture setting, considering the complexity of capture system and convenience for performers.

%\begin{table}[t]\scriptsize
%	\centering
%	\renewcommand{\arraystretch}{1.2}
%	\caption{Distance}
%	\setlength{\tabcolsep}{0.58mm}
%	\begin{tabular}{l|cccc|cccc|cccc}
%	    \hline
%	    \multirow{2}{*}{Inputs} & \multicolumn{4}{c}{Normal(Ours)} &  \multicolumn{4}{|c}{Remote} & \multicolumn{4}{|c}{Extreme} \\ \cline{2-13}
%	    & MPJPE$\downarrow$ & Mesh Err$\downarrow$ & Ang Err$\downarrow$ & Jitter$\downarrow$ & MPJPE $\downarrow$ & Mesh Err$\downarrow$ & Ang Err$\downarrow$ & Jitter$\downarrow$ & MPJPE $\downarrow$ & Mesh Err$\downarrow$ & Ang Err$\downarrow$ & Jitter$\downarrow$\\
%	    \hline 
%	    Lidar Only &65.5&81.5&7.8&0.30 &71.3&92.4&8.3&0.45 &72.8&92.9&7.3&0.40\\
%		5 IMUs only & 70.1&82.6&7.4&0.29&62.7&75.5&7.5&0.45 &50.5&61.7&6.2&0.37\\
%		4-IMU-aid(Ours) & \textbf{58.2} & \textbf{68.9} & \textbf{6.5} &\textbf{0.27}& \textbf{63.8}&\textbf{80.2}&\textbf{6.2}& \textbf{0.41}&\textbf{58.9}&\textbf{72.7}&\textbf{5.8}&\textbf{0.34}\\
%		\hline
%\end{tabular}
%\label{tab:Ablation}
%\end{table}
\begin{figure}[t]
    \centering
    \includegraphics[width=\linewidth]{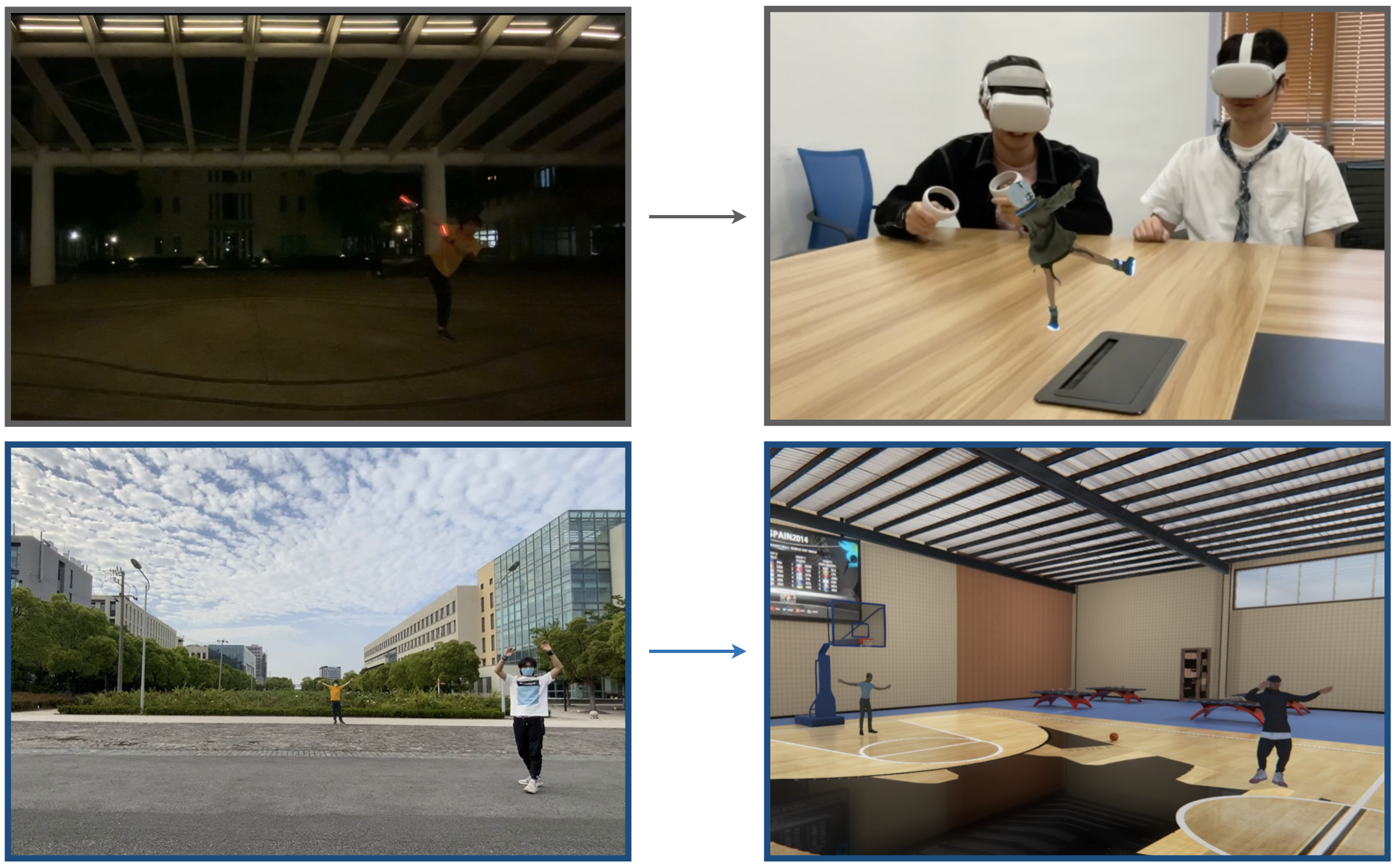}
    \caption{Applications of LIP. It enables conveniently capturing large-scale human performances and driving avatars in various virtual scenes.}
    \label{fig:vrar}
    \vspace{-2ex}
\end{figure}

\subsection{Discussion}
Due to the low frame rate of the utilized LiDAR (10fps), it leads to unsmooth results when capturing extremely fast motions. We plan to improve the prediction modules using or Transformer to interpolate global poses, so as to recover the high-frequency motion details. Moreover, even though we have demonstrated that our method can generalize to different types of LiDAR(e.g. RS-LiDAR-M1~\cite{RS-M1}, OUSTER-0~\cite{OUSTER} and OUSTER-1~\cite{OUSTER}) and IMU sensor(e.g. Xsens Dot~\cite{XSENS}) and Noitom~\cite{noitom}) in this paper, it's still interesting to enhance the capability of our method on handling the domain gaps between more different LiDAR sensors with various point distributions and perception ranges, which relies on further capturing a much larger dataset. It's also promising to extend our pipeline to multi-person or human-object interaction scenes.
\section{Conclusion}
\label{sec:conclusion}
In this paper, we propose a new solution, LiDAR-aid Inertial Poser(LIP), for conveniently capturing human motions in large-scale scenarios, which is occlusion-free and environment-independent. 
As shown in \autoref{fig:vrar}, the recovered human poses and global translations further enable various mix-reality applications like driving avatars in virtual scenes.
% Fig.~\ref{fig:vrar}
%
To explore the new multi-modal setting (light-weight 4 IMUs plus one LiDAR), we design an effective fusion strategy for taking advantage of both LiDAR point cloud and IMU measurements. We also eliminate the translation deviation in a pose-guided manner and gain accurate global trajectories and natural consecutive actions. 
Importantly, we propose a huge LiDAR-IMU hybrid mocap dataset for future research of human motion analysis. 
Extensive experimental results demonstrate the effectiveness of our method and module designs, for convenient and high-quality human motion capture in a large scene.  
We believe our approach and dataset serve as a critical step for light-weight and global human mocap in large-scale scenes, with many potential applications for VR/AR, gaming, filming, or entertainment.

\acknowledgments{
This work was supported by NSFC (No.62206173), Shanghai Sailing Program (No.22YF1428700), and Shanghai Frontiers Science Center of Human-centered Artificial Intelligence (ShangHAI).}

\bibliographystyle{abbrv-doi}

\bibliography{template}
\end{document}